\documentclass{article}
\usepackage{url}
\usepackage{graphicx}
\usepackage{subfigure}

\begin{document}

%
\title{Collaborative Perception for Autonomous Driving: Current Status and Future Trend}
%
%
\author{Shunli Ren \and Siheng Chen \and Wenjun Zhang}
%
%
%

\maketitle              

\begin{abstract}

Perception is one of the crucial module of the autonomous driving system, which has made great progress recently. However, limited ability of individual vehicles results in the bottleneck of improvement of the perception performance. To break through the limits of individual perception, collaborative perception has been proposed which enables vehicles to share information to perceive the environments beyond line-of-sight and field-of-view. In this paper, we provide a review of the related work about the promising collaborative perception technology, including introducing the fundamental concepts, generalizing the collaboration modes and summarizing the key ingredients and applications of collaborative perception. Finally, we discuss the open challenges and issues of this research area and give some potential further directions.

\end{abstract}
%


%
%








\section{Introduction and Motivation}\label{sec:intro}

Autonomous driving is the key technology of intelligent transportation system, and also a very promising engineering project that could fundamentally change the life of human society. Although there has been a lot of progress in both academia and industry in the past decades, autonomous driving is still an important research topic nowadays, especially inspired by the development of computer vision and deep learning recently. 

One of the crucial module of autonomous driving is perception, which targets to perceiving the surrounding environment and extracting information related to navigation, including object detection, tracking, semantic segmentation and so on. Perception used to be regarded as the technique bottleneck of autonomous driving. In the past few years, perception performance has been significantly improved with increasing large-scale training data and developments of deep learning algorithms. However, it is not enough to meet the demand of practical high level of autonomous capacity because high level or fully autonomous vehicles do not require human supervision and totally rely on the perception of vehicles.

\begin{figure}[t]
\centering
\subfigure[\small The occlusion issue.~\cite{wang2020v2vnet}]{
\begin{minipage}[t]{0.5\linewidth}
\centering
\includegraphics[width=1\textwidth]{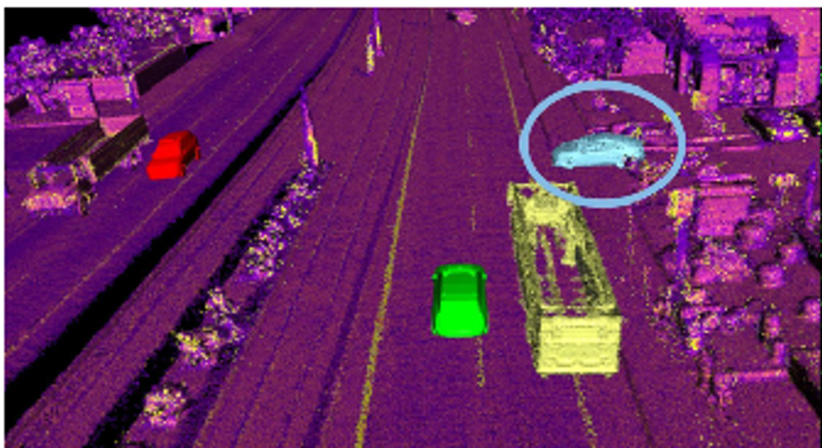}
\end{minipage}%
}%
\subfigure[\small The long range issue.~\cite{Frst2020LRPDLR}]{
\begin{minipage}[t]{0.5\linewidth}
\centering
\includegraphics[width=1\textwidth]{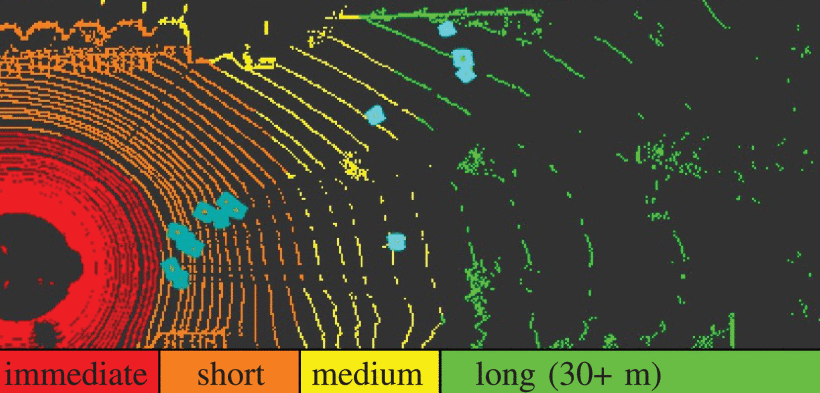}
\end{minipage}%
}%
\centering
\caption{Two issues of individual perception. (a) The occlusion issue: the green vehicle can not see the vehicle in the circle based on its own observation. (b) The long range issue: point cloud becomes sparse as the distance to the ego vehicles increases.}
\label{fig:issues}
\end{figure}

A main factor restricting the perception performance of autonomous driving is that each vehicle perceive surrounding environment based on its own local perception sensors, that is, individual perception. However, individual perception ability is limited and thus perception perception would be decreased by limited field of view, missing modality, sparse sensor data and other negative factors. Two significant issues of individual perception, occlusion and sparse data in long-range, are illustrated in Figure.\ref{fig:issues}. A resolution to these issues is that vehicles in the same area share the collective perception message(CPM) with each other to collaborate to perceive the environments, which is called collaborative perception or cooperative perception.

Benefiting from the better building of communication infrastructure and developing communication technology such as V2X(Vehicle-to-Everything) communication, vehicles could exchange their messages in reliable manners, which enables the collaboration among them. Recent work~\cite{schiegg2021collective,shan2021demonstrations} has demonstrated that collaborative perception among vehicles could improve the accuracy of environmental perception as well as robustness and safety of transportation systems. 

In addition, autonomous driving vehicles are usually equipped with high-fidelity sensors to achieve reliable perception, which causes expensive cost. Collaborative perception can relax the harsh demand of perception equipments of individual vehicles.

Collaborative perception enables autonomous vehicles to overcome some perception limitations such as occlusion and short range of view by sharing information with nearby vehicles and infrastructure. However, achieving real-time and robust collaborative perception requires addressing some challenges caused by communication capacity and noise. Recently, there has been some work to study the strategy of collaboration perception, including what to collaboration, when to collaboration, how to collaboration, alignment of shared information and so on.

In this paper, we review previous work about collaborative perception for autonomous driving. Our contribution is summarized as follows:
i) We induct existing collaboration modes of collaborative perception and respective analysis of methods.
ii) We give a summary of the key ingredients of collaborative perception for autonomous driving and introduce respective research of each ingredient.
iii) We discuss the open challenges and issues in the area of collaborative perception for autonomous driving.

The rest of this paper is organized as follows. In Section.\ref{sec:collaboration mode}, we introduce the pipeline of collaborative perception and its four modes. In Section.\ref{sec:key ind} and~\ref{sec:applications}, we summarize the key ingredients and applications of collaborative perception for autonomous driving respectively. In Section.\ref{sec:trend}, we discuss the open challenges and issues. Finally, Section.\ref{sec:conclusion} concludes this paper.

\section{Collaboration mode: When and What to collaboration}\label{sec:collaboration mode}
The pipeline of perception is that raw data collected by vehicles is firstly fed into encoder, and then the intermediate features output by encoder are decoded to output the final perception results. In this section, we divide the collaboration mode into four categories in terms of when the collaboration occurs in the pipeline, and analyze their advantages and disadvantages in detail respectively.

\begin{figure}[t]
\centering
	\includegraphics[width=0.9\textwidth]{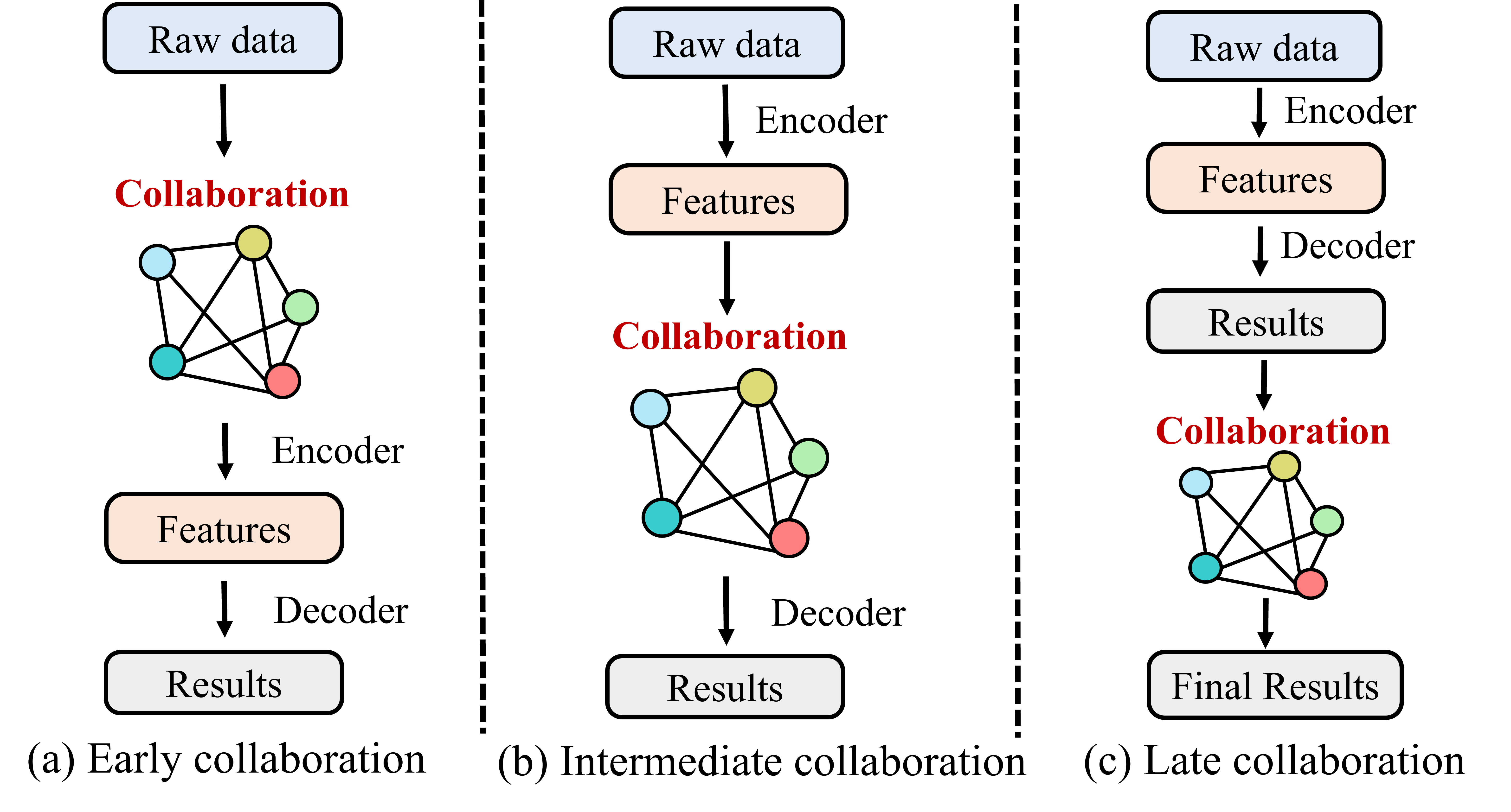}
	\caption{Three collaboration modes.}
	\label{fig:collab mode}
	\vspace{-5mm}
\end{figure}
%

\subsection{Early collaboration}
Early collaboration conducts the collaboration in the input space, which shares raw sensory data among vehicles and infrastructure. It aggregates raw measurements from all vehicles and infrastructure to promote a holistic perspective. Therefore, each vehicle could conduct following processing and finish perception based on the holistic perspective, see Figure.\ref{fig:collab mode}(a), which can fundamentally solve the occlusion and long-range issues occurring in the single-agent perception. \cite{chen2019cooper,arnold2020cooperative} have adopted early collaboration mode and demonstrated its effectiveness with the help of rich information. However, sharing raw sensory data requires a lot of communication and easily congest the communication network with heavy data loads, which impedes its practical usage in most cases.

\subsection{Late collaboration}

Late collaboration conducts the collaboration in the output space, which promotes the fusion of the perception result output by each individual agent to achieve an refinement, see Figure.\ref{fig:collab mode}(c). \cite{miller2020cooperative} adopted late collaboration to develop a perception and localization system and dealt with latency and dropout of the communication link between the two vehicles. \cite{allig2019alignment} studied temporal and spatial alignment of the shared detected objects and proposed to use non-predicted sender state for the transformation and therefore to neglect the sender motion compensation. Although late collaboration is bandwidth-economic, it is very sensitive to the positioning error of the agents and suffers from high estimation errors and noise because of incomplete local observation.

\subsection{Intermediate collaboration}

Intermediate collaboration conducts the collaboration in the intermediate feature space. It enables the transmission of the intermediate features generated by each individual agent's prediction model. After fusion of these features, each agent decodes the fused features and produce the perception results, see Figure.\ref{fig:collab mode}(b). Conceptually, we can squeeze representative information to those features, leading to economic communication bandwidth compared to early collaboration as well as upgraded perception ability compared to late collaboration. A lot of work~\cite{chen2019f,wang2020v2vnet,marvasti2020bandwidth,marvasti2020cooperative,marvasti2020feature} agree with this idea and adopt intermediate collaboration and feature sharing. In practice, the design of this collaboration strategy is algorithmically challenging from two aspects: i) how to select the most beneficial and compact features from the raw measurements for transmission; and ii) how to maximally fuse the other agents' features to enhance each agent's perception ability.

\subsection{Mixed collaboration}
As mentioned above, each collaboration mode has its own advantages and disadvantages. Therefore, some work adopted mixed collaboration which combines two or more collaboration modes to optimize the collaboration strategy. \cite{arnold2020cooperative} proposed to share high level information (late collaboration) where the sensor has high visibility and share low level information (early collaboration) where the visibility is poor. Their method is based on the observation that objects close to a
sensor will have a high density of points and thus are more likely to be detected using a single sensor’s observation. DiscoNet~\cite{yiming2021disconet} leveraged a teacher model employing early collaboration to guide the training of student model employing intermediate collaboration. In reference stage, the communication bandwidth-consuming teacher model is discarded so that student model can keep superior performance with low communication bandwidth because it has learned the knowledge from the teacher model at the training stage.

\section{Key Ingredients of Collaborative Perception Technology}\label{sec:key ind}

\subsection{Collaboration graph}

\begin{figure}[t]
\centering
\subfigure[\small Collaboration graph proposed in \cite{yiming2021disconet}]{
\begin{minipage}[t]{0.3\linewidth}
\centering
\includegraphics[width=1\textwidth]{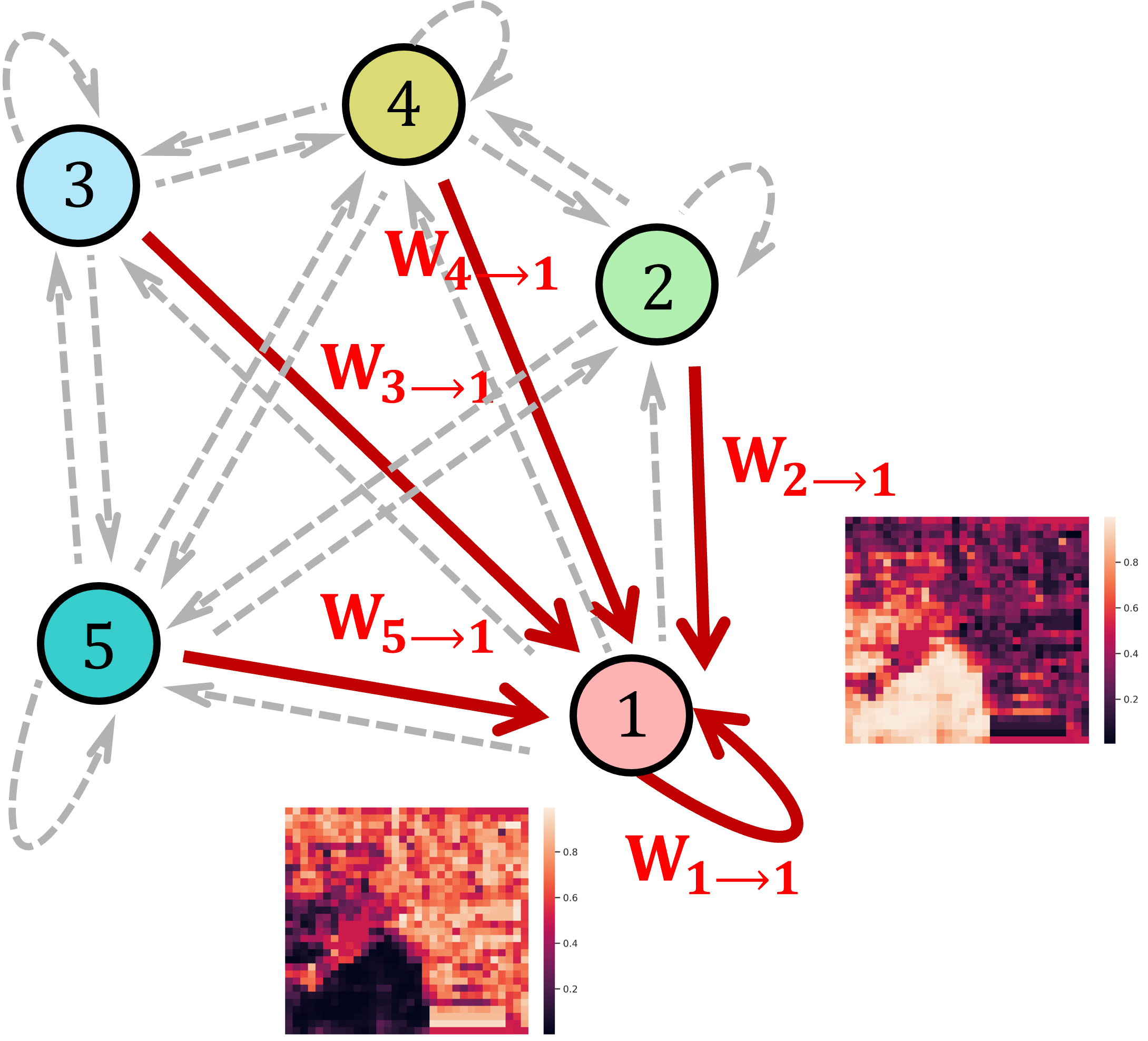}
\end{minipage}%
}%
\subfigure[\small The structure of vehicular networks.~\cite{he2020resource}]{
\begin{minipage}[t]{0.3\linewidth}
\centering
\includegraphics[width=1\textwidth]{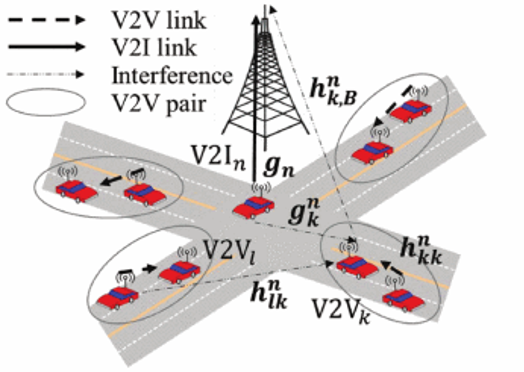}
\end{minipage}%
}%
\subfigure[\small The graph representation of (b).~\cite{he2020resource}]{
\begin{minipage}[t]{0.3\linewidth}
\centering
\includegraphics[width=1\textwidth]{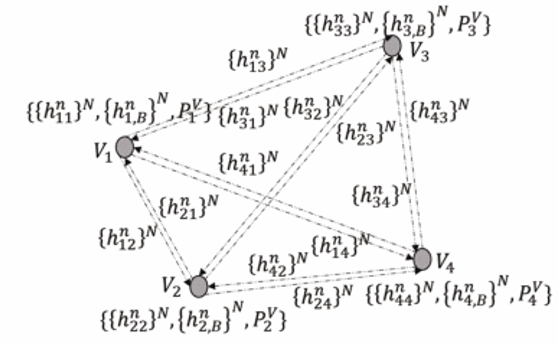}
\end{minipage}
}%
\centering
\caption{Two examples of collaboration graph. (a) Collaboration graph proposed in \cite{yiming2021disconet}: each node represents an agent while collaboration relationship between two vehicles is represented by an edge with a matrix-valued weight which highlights the informative regions. (b)(c) illustrate collaboration graph proposed in \cite{he2020resource}. (b) shows the structure of vehicular networks and (c) is the graph representation of (b). }
\label{fig:collaboration graph}
\end{figure}

Graph is a powerful tool to model collaborative perception process because of its ability to model non-Euclidean data structure and good interpretability. In some work, vehicles participating in collaborative perception compose a complete collaboration graph where each vehicle is a node and the collaboration relationship between two vehicles is the edge between these two nodes. V2VNet~\cite{wang2020v2vnet} leveraged graph neural networks to aggregate and combine messages from other vehicles. \cite{yiming2021disconet} proposed DiscoGraph with matrix-valued edge weight and each element in the matrix reflects the inter-agent attention at a specific spatial region, allowing an agent to adaptively highlight the informative regions, shown in Figure.\ref{fig:collaboration graph}(a). Moreover, some work~\cite{liang2018graph,he2020resource} studied the resource allocation problem in vehicular communication based on graph technology where each vehicle-to-vehicle(V2V) link is considered as a node in the graph, shown in Figure.\ref{fig:collaboration graph}(c).

\subsection{Pose alignment}

Since collaborative perception requires to fuse data from vehicles and infrastructure at different locations and different time, it is significant for successful collaboration to achieve accurate data alignment. \cite{allig2019alignment} presented a review of temporal and spatial alignment including the required coordinate systems and transformations
for cooperative perception. \cite{miller2020cooperative} used extended Kalman filters to compensate for the position and relative distance measurements by extrapolation under the latency consideration. V2VNet~\cite{wang2020v2vnet} and DiscoNet~\cite{yiming2021disconet} adopted the pose-aware strategy whose assumption is that vehicles have access to accurate poses and locations of itself and its collaborators so that collaborative perception can success by learned spatially-aware feature maps fusion. \cite{vadivelu2020learning} pointed that localization noise is common in the real world and performance of pose-aware strategy degrades below single-agent performance under realistic amounts of noise. They proposed end-to-end learnable neural reasoning layers that learn to estimate pose errors to make vehicles reach a consensus about those errors. 
\cite{glaser2021overcoming} leveraged neural layers to learn the data correspondence without the requirements of pose information of other agents.

\subsection{Information fusion}

Information fusion is a core component in multi-agent systems, which targets to fuse the most informative part from other agents in an effective manner. CommNet~\cite{Sukhbaatar2016LearningMC} adopted the averaging operation to conduct information fusion and VAIN~\cite{Hoshen2017VAINAM} considered an attention mechanism to determine which agents would share information with. Almost all later fusion methods~\cite{Jiang2018LearningAC,liu2020who2com,liu2020when2com} adopted attention mechanism since it can adaptively compute the relationship between two agents. DiscoNet~\cite{yiming2021disconet} leveraged a mask to reflect inter-agent attention at each spatial region, and compare its performance with some basic fusion methods such as summation, average, maximization, concatenation and state-of-the-art methods.

\subsection{Resource allocation with reinforcement learning}

Limited communication bandwidth in realistic environments asks us to fully use the available communication resources, which  makes resource allocation and spectrum sharing significant. In vehicular communication environment, fast changing channel conditions and increasingly service requirements make the optimization of the allocation problem very complex and it hard to use conventional optimization methods to solve it. Some work leverage multi-agent reinforcement learning(MARL) to solve the optimization problem. \cite{Aoki2020CooperativePW} used deep reinforcement learning to select the data to transmit and mitigated the network load. With MARL, \cite{liang2018graph,he2020resource} focus on optimize resources allocation and \cite{Liang2019SpectrumSI} focused on spectrum sharing. \cite{AbdelAziz2020VehicularCP} introduced federated reinforcement learning to speed up the training process.

\section{Applications of Collaborative Perception}\label{sec:applications}

Collaborative perception can be applied to a lot of tasks involving multi-agent perception. In this section, we focus two important tasks for autonomous driving and group intelligence, point clouds based 3D object detection and semantic segmentation of 3D scenes, to which collaborative perception is applied.

\subsection{Collaborative 3D object detection}

\begin{figure}[t]
\centering
\subfigure[\small Collaborative object detection.~\cite{yiming2021disconet}]{
\begin{minipage}[t]{0.5\linewidth}
\centering
\includegraphics[width=1\textwidth]{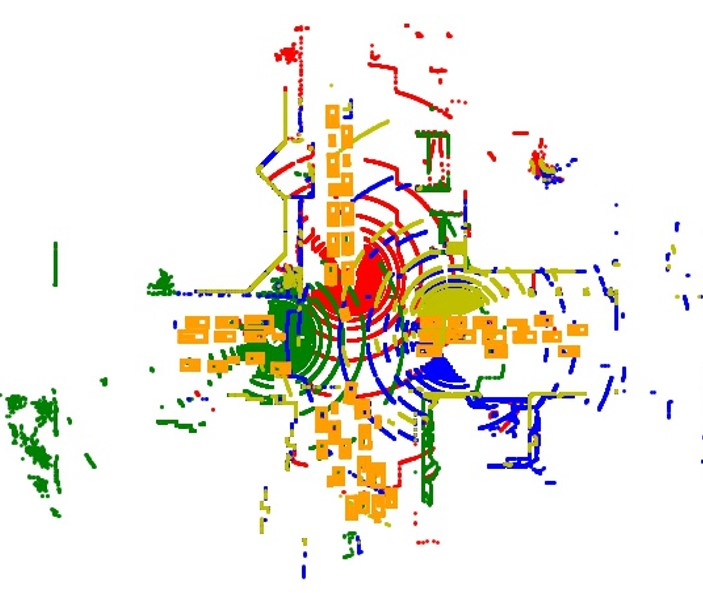}
\end{minipage}%
}%
\subfigure[\small Collaborative semantic segmentation.~\cite{glaser2021overcoming}]{
\begin{minipage}[t]{0.5\linewidth}
\centering
\includegraphics[width=1\textwidth]{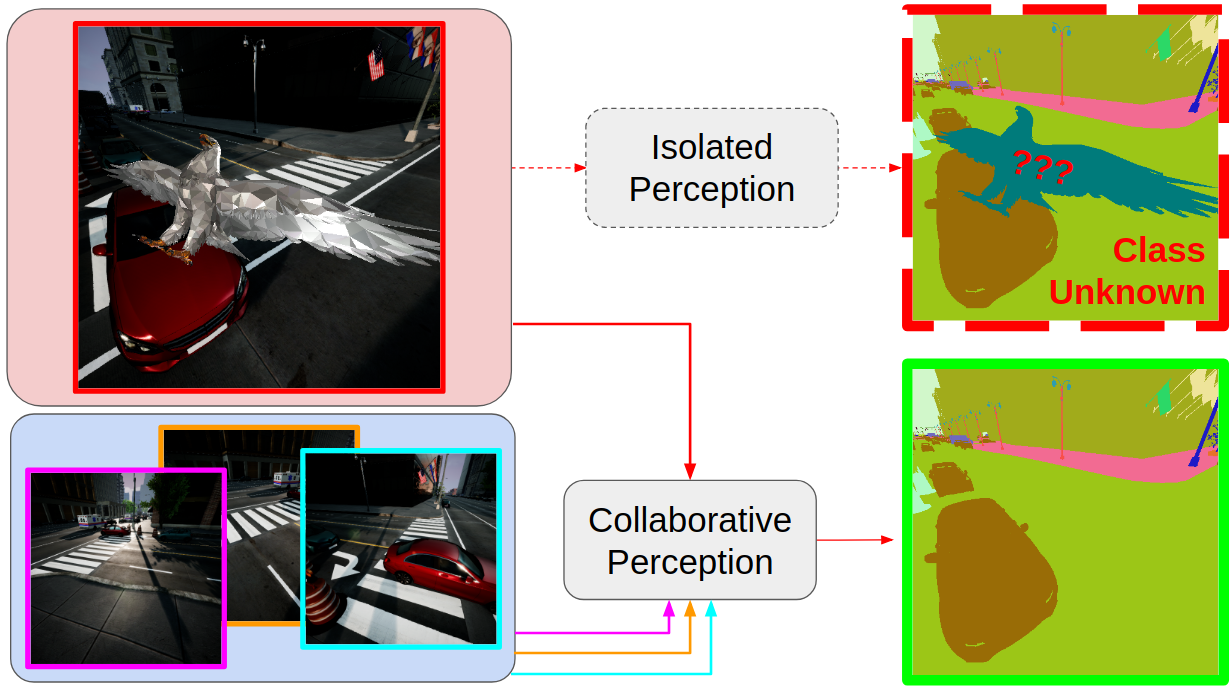}
\end{minipage}%
}%
\centering
\caption{Two task where collaborative perception is applied. (a) An example of collaborative 3D object detection where vehicles with observations in different colors aim to detect the surrounding vehicles in a collaborative manner. (b) In the
presence of an unexpected obstruction, collaborative semantic segmentation gives a desirable output while individual perception may fail.}
\label{fig:tasks}
\end{figure}

\begin{figure}[t]
\centering
\subfigure[\small Ego-vehicle]{
\begin{minipage}[t]{0.3\linewidth}
\centering
\includegraphics[width=1\textwidth]{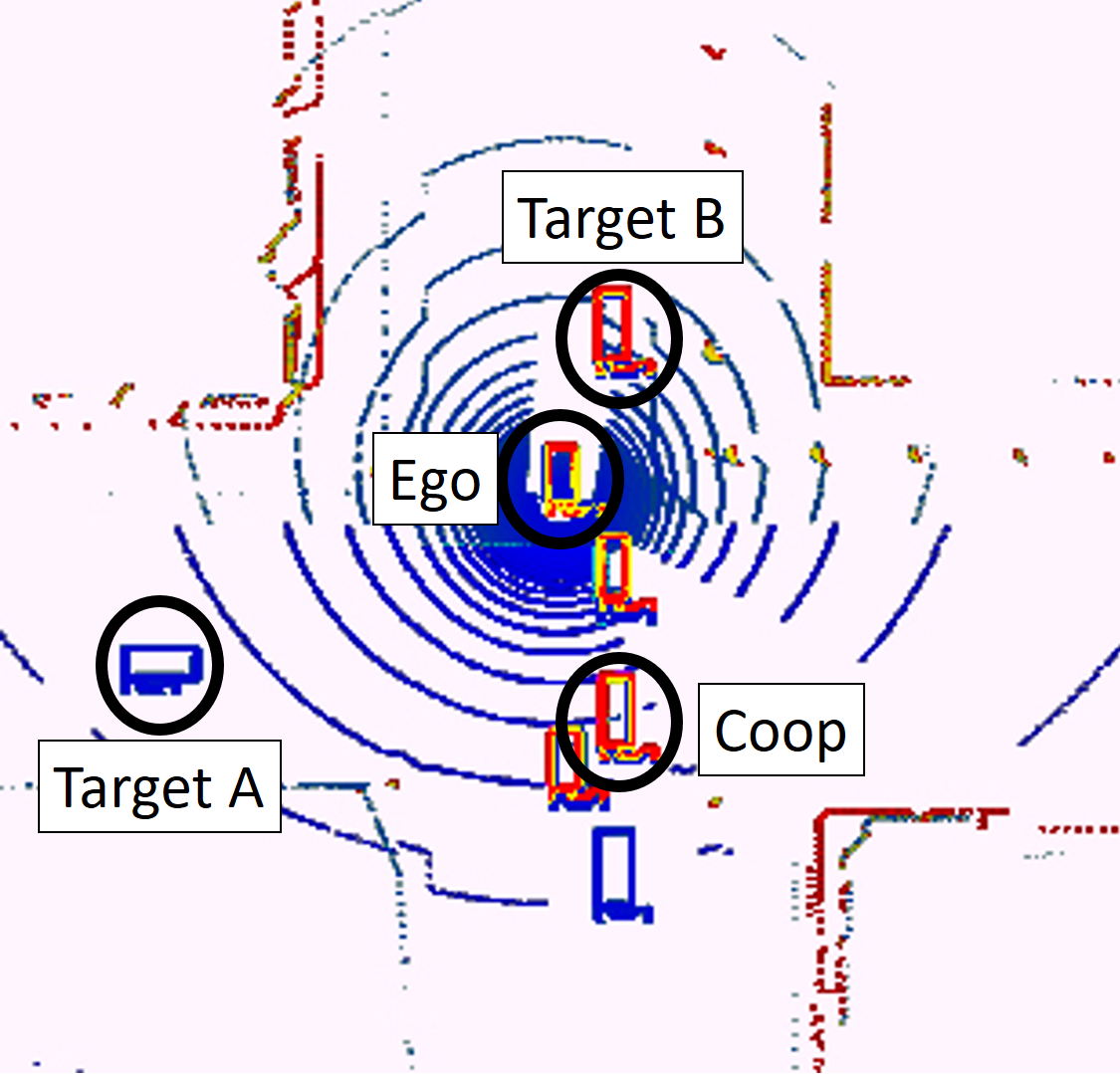}
\end{minipage}%
}%
\subfigure[\small Coop-vehicle]{
\begin{minipage}[t]{0.3\linewidth}
\centering
\includegraphics[width=1\textwidth]{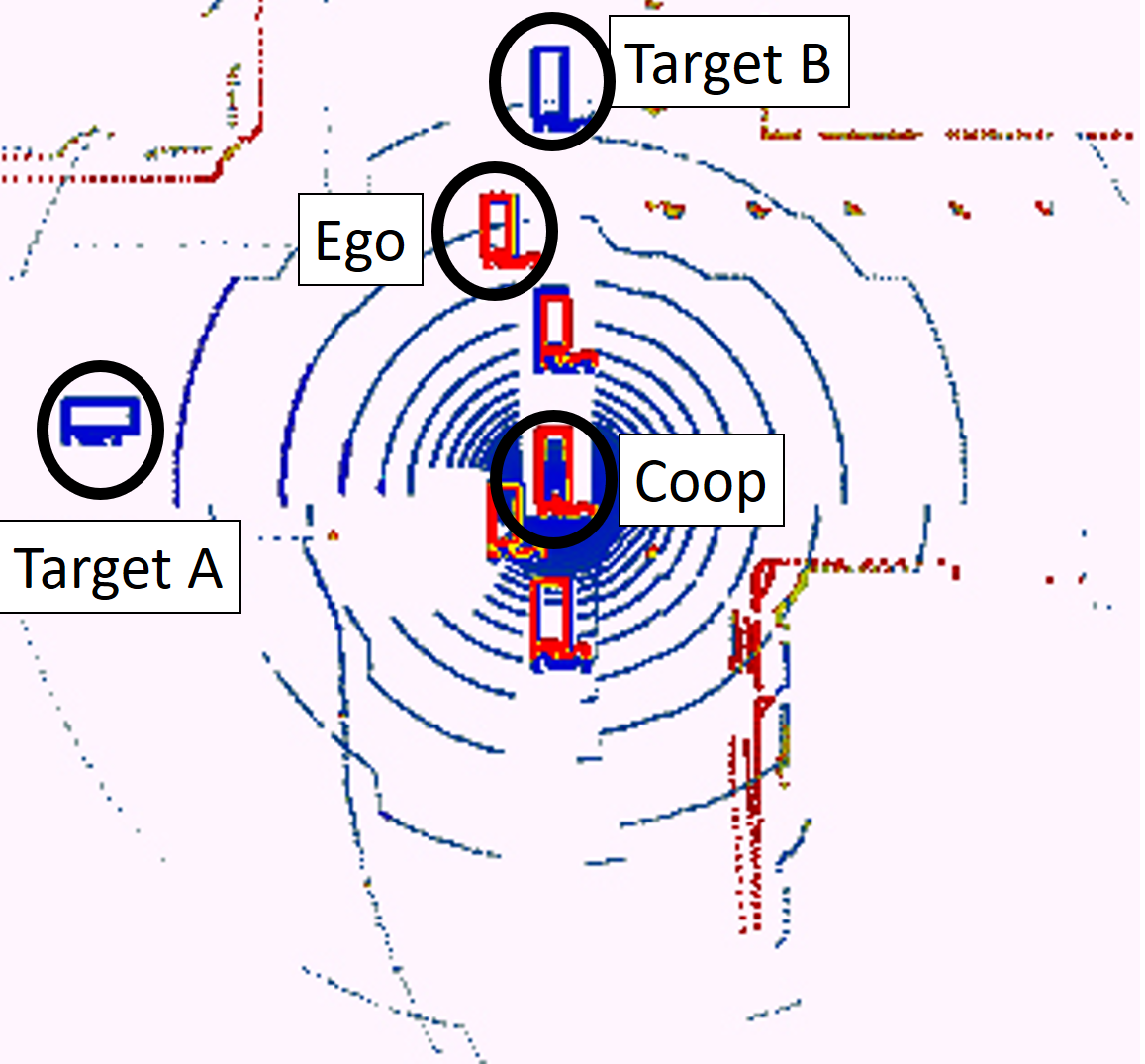}
\end{minipage}%
}%
\subfigure[\small Collaboration]{
\begin{minipage}[t]{0.3\linewidth}
\centering
\includegraphics[width=1\textwidth]{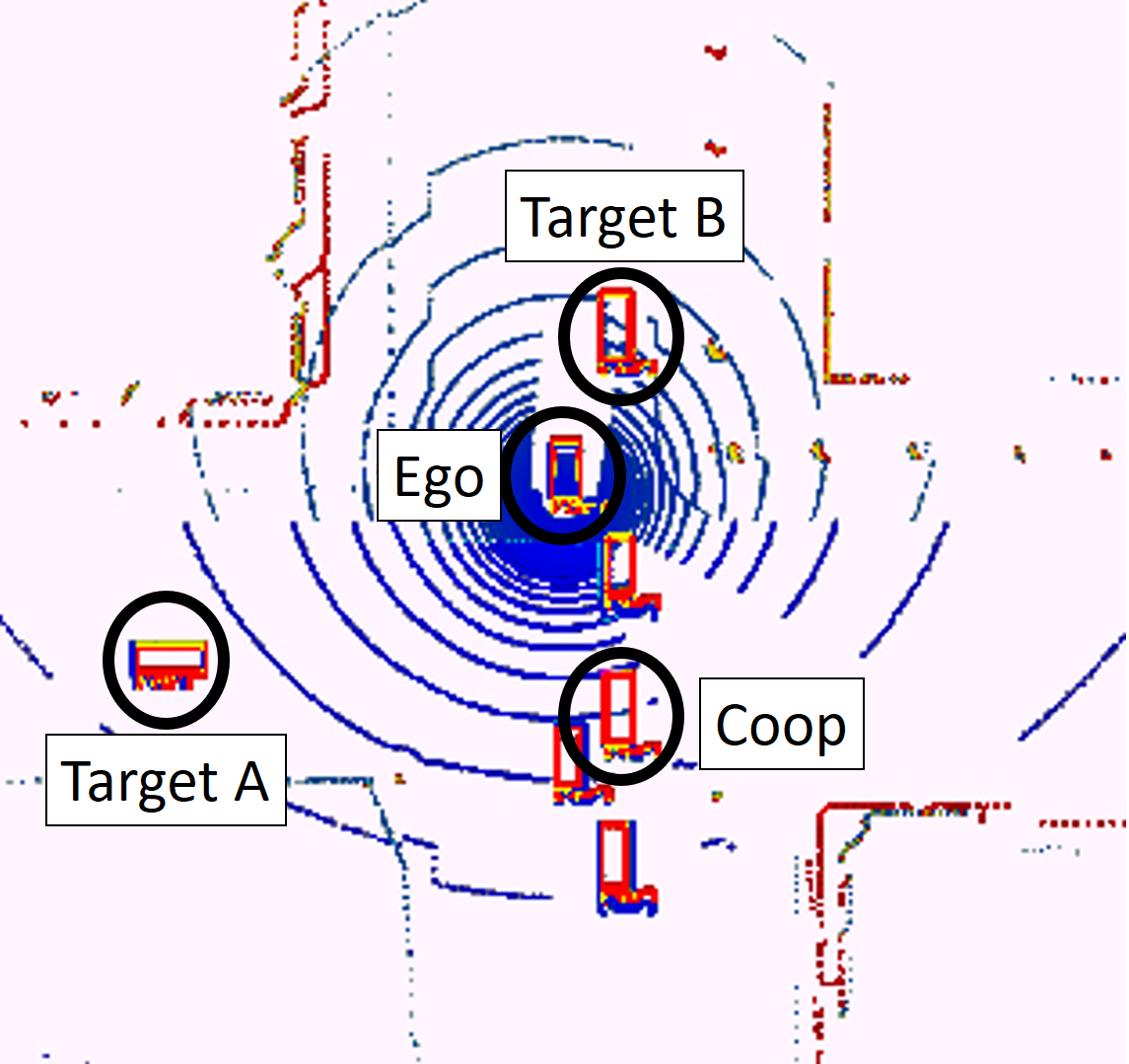}
\end{minipage}
}%
\centering
\caption{Comparison between the performance of single-vehicle object detection and collaborative object detection based on 3D LIDAR point clouds. Red and blue bounding boxes represent the output of the object detection and ground truth; (a) Ego-vehicle single view object detection, (b) Coop-vehicle single view object detection and (c) The collaborative object detection at ego-vehicle. Cited from~\cite{marvasti2020cooperative}}
\label{fig:results comparison}
\end{figure}

LIDAR-point-cloud based 3D object detection attracts most attention in the study of collaborative perception. The reason is as follows: i) LiDAR point clouds have more spatial dimension than images and video. ii) LIDAR point clouds can preserve private information in some extent such as people's faces and license plate numbers. iii) point data is a proper data type for fusion because there is less loss when point data are aligned from different poses than pixels. iv) 3D object detection is an essential task for autonomous driving perception, on which many tasks such as tracking and motion prediction are based.

Figure.\ref{fig:tasks}(a) shows an example of collaborative object detection where vehicles with observations in different colors aim to detect the surrounding vehicles in a collaborative manner. Figure.\ref{fig:results comparison} compares 3d object detection results of single perception and collaborative perception based point clouds, which illustrates the effectiveness of collaboration perception.

\subsection{Collaborative semantic segmentation of 3D scenes}

Semantic segmentation of 3D scenes is also a key task required for autonomous driving. Collaborative semantic segmentation of 3D scenes targets to produce semantic segmentation masks for each agent given observations (images, LIDAR point clouds, \textit{etc.}) of 3D scenes from several agents. \cite{liu2020who2com,liu2020when2com,glaser2021overcoming} focus on collaborative semantic segmentation with limited communication bandwidth. Figure.\ref{fig:tasks}(b) illustrates the effectiveness of collaborative semantic segmentation leveraging multi views in the presence of an unexpected obstruction.

\section{Open Challenges and Issues}\label{sec:trend}
In this section, we will discuss existing open challenges and issues of collaborative perception and give some potential further directions.
%

\subsection{Communication Robustness}
Efficient collaboration depends on reliable communication among agents. However, communication is not perfect in practice: i) As the number of vehicles in the network increases, available communication bandwidth for each vehicle is limited; ii) It is hard for vehicles to receive real-time information from other vehicles due to inevitable communication delay; iii) Communication is likely to be interrupted sometimes, resulting in communication drops; iv) V2X communication suffers from attacks~\cite{Tu_2021_ICCV} and cannot always provide reliable service. Although communication technologies is keep developing and improving  quality of communication service, the problems mentioned above will still exist for a long time. However, most existing work adopt the assumption that information can be sharing in a real time and lossless manner, so it is significant for further work to consider those communication constrains and design robust collaborative perception systems.


\subsection{Heterogeneous and cross-modality}

Most collaborative perception work pay attention to LIDAR point clouds based perception. However, there are more types of data useful for perception, such as images and mmWave radar points. It is a potential approach for more effective collaboration to leverage the multi-modal sensory data. In addition, there are different levels of autonomous vehicles which provides information with different quality in some scenes. Therefore, how to collaborate in the heterogeneous vehicles network is an issue for further practical application of collaborative perception. Unfortunately, few work focus on heterogeneous and cross-modal collaborative perception, making it a open challenge.

\subsection{Large-scale dataset}

As mentioned in Section.\ref{sec:intro}, perception performance has been boosted by increasing large-scale dataset and development of deep learning methods. However, existing datasets in the research area of collaborative perception are either small in scale or not public. For example,~\cite{chen2019cooper} regards the ego vehicle at different timestamps as multiple collaborator vehicles in KITTI~\cite{geiger2013vision}, and V2V-Sim generalized based on a high-fidelity LIDAR simulator~\cite{manivasagam2020lidarsim} proposed by V2VNet~\cite{wang2020v2vnet} is not publicly available. Recently, ~\cite{DAIR-V2X2021} was released as a large scale dataset for Vehicle-Infrastructure collaborative autonomous driving. However, it does not include V2V scenario and is not public currently. Consequently, the absence of public large-scale dataset preclude further development of collaborative perception. In addition, most dataset is based on simulation. Although simulation is a economic and safe approach to validate the algorithm, realistic dataset is in demand to make collaborative perception applied in practice.

\section{Conclusion}\label{sec:conclusion}

This review of collaborative perception has covered the major technical details and applications for autonomous driving. We introduce the concept of collaborative perception and provide an analysis of the advantages and disadvantages of different collaboration modes. We then induct the key ingredients and two important application tasks of collaborative perception technology. Finally, we discuss the open challenges and issues in this research area and give some potential further directions of collaborative perception.

\textbf{Acknowledgment.} This research is partially supported by the Science and Technology Commission of Shanghai Municipal under Grant 21511100900.

\bibliographystyle{unsrt}
\bibliography{main}

\end{document}